\pdfoutput=1

\documentclass[11pt]{article}

\usepackage{ACL2023}

\usepackage{supertabular}
\usepackage{longtable}

\usepackage{amsmath}

\usepackage{times}
\usepackage{latexsym}
\usepackage[nameinlink]{cleveref}

\usepackage{algorithm}
\usepackage{algorithmicx}
\usepackage{algpseudocode}
\usepackage{todonotes}
\usepackage{soul} 
\usepackage{mdframed} 

\usepackage[T1]{fontenc}

\usepackage{graphicx}

\usepackage[utf8]{inputenc}

\usepackage{microtype}

\usepackage{inconsolata}

\usepackage{ dsfont }

\newcommand{\E}{\mathds{E}}

\renewcommand{\phi}{\ensuremath{\varphi}}

\title{Legal Extractive Summarization of U.S. Court Opinions}

\author{Emmaneul Bauer \\
  ETH Zurich \\
  \texttt{bauerem@student.ethz.ch} \\\And
  Dominik Stammbach \\
  ETH Zurich \\
  \texttt{dominsta@ethz.ch} \\\And
  Nianlong Gu \\
  ETH Zurich \\
  \texttt{nianlong@ini.ethz.ch} \\\And
  Elliott Ash \\
  ETH Zurich \\
  \texttt{ashe@ethz.ch} \\}
\begin{document}


\maketitle
\begin{abstract}
This paper tackles the task of legal extractive summarization using a dataset of 430K U.S. court opinions with key passages annotated. According to automated summary quality metrics, the reinforcement-learning-based MemSum model is best and even out-performs transformer-based models. In turn, expert human evaluation shows that MemSum summaries effectively capture the key points of lengthy court opinions. Motivated by these results, we open-source our models to the general public. This represents progress towards democratizing law and making U.S. court opinions more accessible to the general public.

\end{abstract}

\section{Introduction}

How many people have argued about \textit{Dobbs v. Jackson}, the 2022 U.S. Supreme Court abortion case, without having read the associated judicial opinions? Even among trained attorneys, it is not trivial to read a set of opinions spanning 213 pages. While many news outlets have provided more accessible summaries of the majority opinion, it is not easy to find a decent summary of the dissenting opinion. And countless other judicial opinions in U.S. courts have never been put into a more digestible format. 

A promise of recently developed AI technologies for natural language processing is to automatically analyze and summarize technical documents, such as judicial opinions. This paper focuses on the task of \textit{extractive summarization} -- that is, extracting key passages from a long document that concisely summarize its main points. We use a labeled dataset of 430K judicial opinions from U.S. courts to train a set of neural-net summarizers that try to reproduce the human labels. In a held-out test set of unseen documents, a reinforcement-learning-based architecture called MemSum out-performs other baselines, including a long-context transformer architecture. 

Qualitatively, the produced summaries are of extremely high quality. Consider for example the summaries for \textit{Dobbs}, shown in Table \ref{tab:summaries_dobbs_v_jackson}. For both the majority (top panel) and dissent (bottom panel), the key legal and policy arguments are effectively extracted and reproduced, with a 100$\times$ compression ratio in terms of document length. In a more extensive blinded human validation task where a lawyer-annotator compared human summaries to machine summaries, we further substantiate the quality of the machine-generated summaries.  

Motivated by these evaluations, we open-source our models on github.\footnote{\url{https://github.com/bauerem/legal_memsum}} Our models will now be available for use by the legal community and the broader public.  

On the academic side, these contributions add to the applied NLP field on legal document summarization \cite{jain-2021-les}. The early work in this area used a combination of rule-based heuristics and unsupervised clustering approaches \citep[among others][]{MOENS1997727, farzindar-2004-letsum, hachey-2006-holj, yousfimonod-2010-ml4sld, kim-etal-2012-holj}, on a range of small-scale datasets in various countries and domains. 
The closest recent work on extractive summarization of U.S. legal documents is \citet{kornilova-eidelman-2019-billsum}, who provide a new dataset of legislation with annotated summaries, BillSum, and provide some extractive summary baselines \cite[see also][]{jain-etal-2021-cawesumm}. We produce complementary baselines for U.S. court opinions, via experimenting with two different dedicated long-document summarization approaches.

\begin{table*}[]
    \caption{Extractive Summaries: \textit{Dobbs v Jackson}'s Majority and Dissenting Opinions}
    \centering
    \tiny
    \begin{tabular}{c  c | p{13.5cm}}
    \multicolumn{3}{c}{\textbf{Majority Opinion by Justice Alito (915 sentences in original)}} \\
Rank & Position & Extracted sentence \\ \hline
7 & 77 &  That provision has been held to guarantee some rights that are not mentioned in the Constitution, but any such right must be “deeply rooted in this Nation’s history and tradition” and “implicit in the concept of ordered liberty."\\
8 & 99 &  “Except in a medical emergency or in the case of a severe fetal abnormality, a person shall not intentionally or knowingly perform... or induce an abortion of an unborn human being if the probable gestational age of the unborn human being has been determined to be greater than fifteen (15) weeks.\\
5 & 150 &  Finally, we consider whether a right to obtain an abortion is part of a broader entrenched right that is supported by other precedents.\\
1 & 155 &  The Constitution makes no express reference to a right to obtain an abortion, and therefore those who claim that it protects such a right must show that the right is somehow implicit in the constitutional text.\\
2 & 182 &   The regulation of a medical procedure that only one sex can undergo does not trigger heightened constitutional scrutiny unless the regulation is a “mere pretex[t] designed to effect an invidious discrimination against members of one sex or the other.” \\
3 & 185 &  And as the Court has stated, the “goal of preventing abortion” does not constitute “invidiously discriminatory animus” against women.\\
4 & 187 &  Accordingly, laws regulating or prohibiting abortion are not subject to heightened scrutiny.\\
6 & 188 &  Rather, they are governed by the same standard of review as other health and safety measures.\\ \\
\end{tabular}

    \begin{tabular}{c  c | p{13.5cm}}
    \multicolumn{3}{c}{\textbf{Dissenting Opinion by Justices Breyer, Kagan, and Sotomayor (1,128 sentences in original)}} \\
Rank & Position & Extracted sentence \\ \hline
10 & 60 &  As of today, this Court holds, a State can always force a woman to give birth, prohibiting even the earliest abortions.\\
3 & 91 &  \textit{Stare decisis} is the Latin phrase for a foundation stone of the rule of law: that things decided should stay decided unless there is a very good reason for change.\\
4 & 92 &  It is a doctrine of judicial modesty and humility.\\
8 & 93 & Those qualities are not evident in today’s opinion.\\
2 & 104 &  \textit{Stare decisis}, this Court has often said, “contributes to the actual and perceived integrity of the judicial process” by ensuring that decisions are “founded in the law rather than in the proclivities of individuals.'' \\
6 & 123 &  We believe in a Constitution that puts some issues off limits to majority rule.\\
7 & 124 &  Even in the face of public opposition, we uphold the right of individuals—yes, including women—to make their own choices and chart their own futures.\\
9 & 160 &  Recognizing that “arguments [against \textit{Roe}] continue to be made,” we responded that the doctrine of \textit{stare decisis} “demands respect in a society governed by the rule of law.''\\
5 & 163 &  And we avowed that the “vitality” of “constitutional principles cannot be allowed to yield simply because of disagreement with them.''\\
1 & 182 &  “It is settled now,” the Court said—though it was not always so—that “the Constitution places limits on a State’s right to interfere with a person’s most basic decisions about family and parenthood, as well as bodily integrity.''\\
\end{tabular}

   \flushleft {\footnotesize \textit{Notes.} MemSum's extracted sentences from the majority opinion (top panel) and dissenting opinion (bottom panel) published in the 2022 U.S. Supreme Court case \textit{Dobbs v. Jackson}. \textit{Rank} means the order of extraction by MemSum, which is a proxy for the sentence's importance (lower number = more important). \textit{Position} is the ordinal position of the sentence in the original opinion's sequence. \textit{Extracted sentence} is the sentence text.}
    \label{tab:summaries_dobbs_v_jackson}
\end{table*}

Legal document summarization is part of the broader literature on legal NLP, which has seen rapid progress in recent years \citep[see e.g.,][]{zhong-etal-2020-nlp, chalkidis-etal-2020-legal, peric_et_al_legal_LM, chalkidis-etal-2022-lexglue}. Many legal tasks are natural-language tasks, and massive legal corpora have become increasingly available for computational analysis \cite[e.g.][]{henderson2022pile}. An interesting challenge in the legal domain is the relatively long document length \cite{chalkidis-etal-2022-lexglue, legalBERT}, which poses unique challenges for a range of tasks \cite[e.g.][]{long_document_summarization, jain-2021-les}. Our solution to that problem in the case of summarization -- a reinforcement-learning-based approach that easily scales to hundreds or even thousands of sentences -- could be useful to legal NLP more generally.

In terms of applied NLP methodology, our research is useful in showing a context where a relatively light-weight reinforcement-learning-based model (\citealt{MemSum}; see also \citealt{luo-etal-2019-reading, narayan-etal-2018-ranking, robust_deep_RL_extractive_legal_summarization}) out-performs a more heavy-duty transformer-based model. One of the original motivations for approaches like MemSum is that they can scale to longer documents than BERT-based models. Now that long-context transformers like LongFormer have relaxed the document-length constraint, we might expect those models to out-perform MemSum. At least in our setting, they do not.

Beyond its place in the academic literature, we hope that our models will help democratize legal practice by making the primary sources -- i.e., judicial opinions -- more accessible to attorneys, to policymakers, and to the broader public \citep{carroll2006movement}. Given recent work showing that judges use Wikipedia articles about precedents in their opinions \citep{thompson2022trial}, there is proven value in making such models available. 

\section{Methods} 

\subsection{Data}

We have data on 436,889 judicial opinions published by courts in the United States (both state and federal) from the years 1755 through 2016. Each document (a majority opinion) is accompanied by human annotations of key passages, which we take as an extractive summary. These annotations are used by lawyers to understand the distinctive features of a case and the most relevant points of law made by the associated ruling. The average opinion contains 86 sentences, while the  average summary contains 6 sentences, making for an average compression ratio of 15.8\% (Appendix Table \ref{tab:dataset_stats}). 

We train the model on most of the data -- 410,732 opinions -- with 13,146 opinions used for validation and 13,011 held out for test evaluation. 

\subsection{Quantitative Results}


\begin{table*}[ht!]
\footnotesize
\caption{Extractive Summarization Model Performance}
\centering 
\begin{tabular}{l c c c}
Model & \small ROUGE-1 & \small ROUGE-2 & \small ROUGE-L \\ \hline
Lead-10 & 30.5 & 12.0 & 26.7 \\  \hline
LongFormer  & 54.0 & 46.7 & 39.9  \\
LawFormer  & \underline{56.0} & \underline{48.4}  & \underline{41.1} \\
MemSum & \textbf{62.8} & \textbf{55.3} & \textbf{61.1} \\ \hline
Oracle & 85.5 & 80.2 & 84.5 \\ \hline
\end{tabular}

\flushleft  {\footnotesize  \textit{Notes.} F1 scores (in \%) on the test set, with models indicated by row and ROUGE variants indicated by column. Best scores by column in bold; second-best underlined.}
\label{tab:rouge}
\end{table*}

\subsection{Summarization Models}

The average opinion in our dataset contains 2'745 words (Appendix Table \ref{tab:dataset_stats}), which translates to 3,500 tokens, about seven times the maximum sequence length (512 tokens) of typical transformer models such as BERT \cite{devlin-2018-bert}. Further, about 27\% of opinion documents are longer than 4096 tokens, the maximum sequence length of standard long-document transformers using sparse attention \cite{beltagy-2020-longformer, bigbird}. Our summarization modeling choices have to address this issue of unusually long documents \cite{long_document_summarization}.  We experiment with two types of models: (1) a neural architecture built to extract sentences from long documents using reinforcement learning learning, and (2) an extra-long-document transformer encoder which can handle input up to sequence length of 16'834 tokens.

\paragraph{MemSum.}

The first extractive summarizer model we use is MemSum -- ``Multi-step Episodic Markov decision process extractive SUMmarizer'' \cite{MemSum}. This model uses reinforcement learning to train a policy network to iteratively select sentences from text. At each extraction step, the network considers the local context of each candidate sentence (encoded by bi-LSTM), the global context of the whole document (also encoded by bi-LSTM), and the historical context of previously extracted sentences (encoded by a mini-transformer). The extractor (a feedforward network) takes these concatenated encodings as input and decides which sentence to add, or else to stop extracting. The model learns by rewarding the agent for a higher ROUGE score (which is not differentiable, hence the use of reinforcement learning). See Appendix \ref{app:memsum} for more details.

\paragraph{LongFormer Encoder.}

The second model uses a sparse-attention transformer architecture based on the LongFormer encoder from \citet{beltagy-2020-longformer}. While the original LongFormer accommodates sequence lengths up to 4,096 tokens, this variant handles sequence lengths up to 16'384 tokens -- enough for almost all court opinions in our dataset. Similar to the discriminator objective in \citet{clark2020electra}, we implement extractive summarization as a binary token-level prediction task -- that is, to predict for each token whether it is part of the extractive summary or not. At inference time, a sentence is selected as part of the summary if the average score across a sentence's tokens is greater than 0.5. 

We present two flavors of this approach. First, "LongFormer'' is the long-context version of \citet{beltagy-2020-longformer} fine-tuned on the token prediction task using our dataset. Second,``LawFormer'' is similar, but also pre-trained on a legal BART objective \citep{lewis2019bart} using 6 million U.S. court opinions (see Appendix 
\ref{app:lawformer} for details).

\subsection{Evaluation}

Models are assessed by the textual overlap of the machine-selected summary sentences with the human-selected gold summary sentences in the held-out test set. This overlap is computed by ROUGE scores, with ROUGE-1 measuring unigram overlap, ROUGE-2 measuring bigram overlap, and ROUGE-L measuring the longest common subsequence. We report ROUGE F1 scores, computed as the harmonic mean of ROUGE recall (how much of the gold summary features are covered by the machine summary) and ROUGE precision (how much of the machine summary features are contained in the gold summary). 

As an upper bound on performance, we compute these ROUGE scores for the sentences extracted by an ``Oracle'' model that approximates scores for the ground truth. See Appendix \ref{sec:oracle_details} for more details. 
 
\section{Results}

The main results for summarizer model performance are reported in Table \ref{tab:rouge}. The columns indicate the different ROUGE score variants (unigram, bigram, and longest subsequence), while the rows indicate the models/baselines. The cell values are ROUGE F1 scores reported in \% (0-100).

The first row, Lead-10, is a lower-bound baseline where the extracted summary is the first ten sentences of the opinion. The bottom row, Oracle, is the upper-bound baseline proxying for the ground truth. As desired, these rows respectively give the lowest and highest values for each column. 

The second row reports results for LongFormer. That model achieves a substantial increase over the Lead-10 baseline in all metrics, including a 4x increase for ROUGE-2 F1. Looking to LawFormer in row 3, we see that further pre-training on U.S. court opinions does help but only marginally. 

Finally in row 4, we consider the MemSum model. It achieves the best overall results. MemSum consistently improves over the other models by at least 6\% in every metric measured. That includes an increase of 1.5x in ROUGE-L over the second best model. Hence, for our qualitative validation and open-sourced summaries, we use MemSum.


\subsection{Qualitative Results}

As already discussed in the introduction, an initial sense of the summarizer's quality can be seen in  Table \ref{tab:summaries_dobbs_v_jackson}'s summaries of the opinions in \textit{Dobbs}. The majority opinion (top panel) is compressed from 915 sentences to 8 sentences, while the dissenting opinion (bottom panel) is compressed from 1,128 sentences to 10 sentences. The first column, Rank, indicates the ordering in which MemSum extracts sentences, and can be used as a rough ranking of which sentences are most important in an opinion. 

To generalize this qualitative validation, we used a list of 25 U.S. Supreme Court opinions that had been cited most often by law review articles. Of these notable opinions, 14 were not part of the model's training set. For this subset, the generated summaries are listed in Appendix \ref{app:qual}. We feel that overall, the summaries are of high quality. 

Next, we arranged blind human annotation by a trained U.S. lawyer who was familiar with these 14 cases. The lawyer observed the ground-truth (human-generated) summaries and the machine-generated summaries, presented in random order without labels. For each case, the annotator ranked the pair on a five-point scale: A clearly better than B (-2) , A slightly better than B (-1) , could not decide (0), B slightly better than A (+1), and B clearly better than A (+2). 

\begin{figure}
    \centering
    \caption{Human Validation: Relative Quality of Machine Summaries}
    \includegraphics[width=\columnwidth]{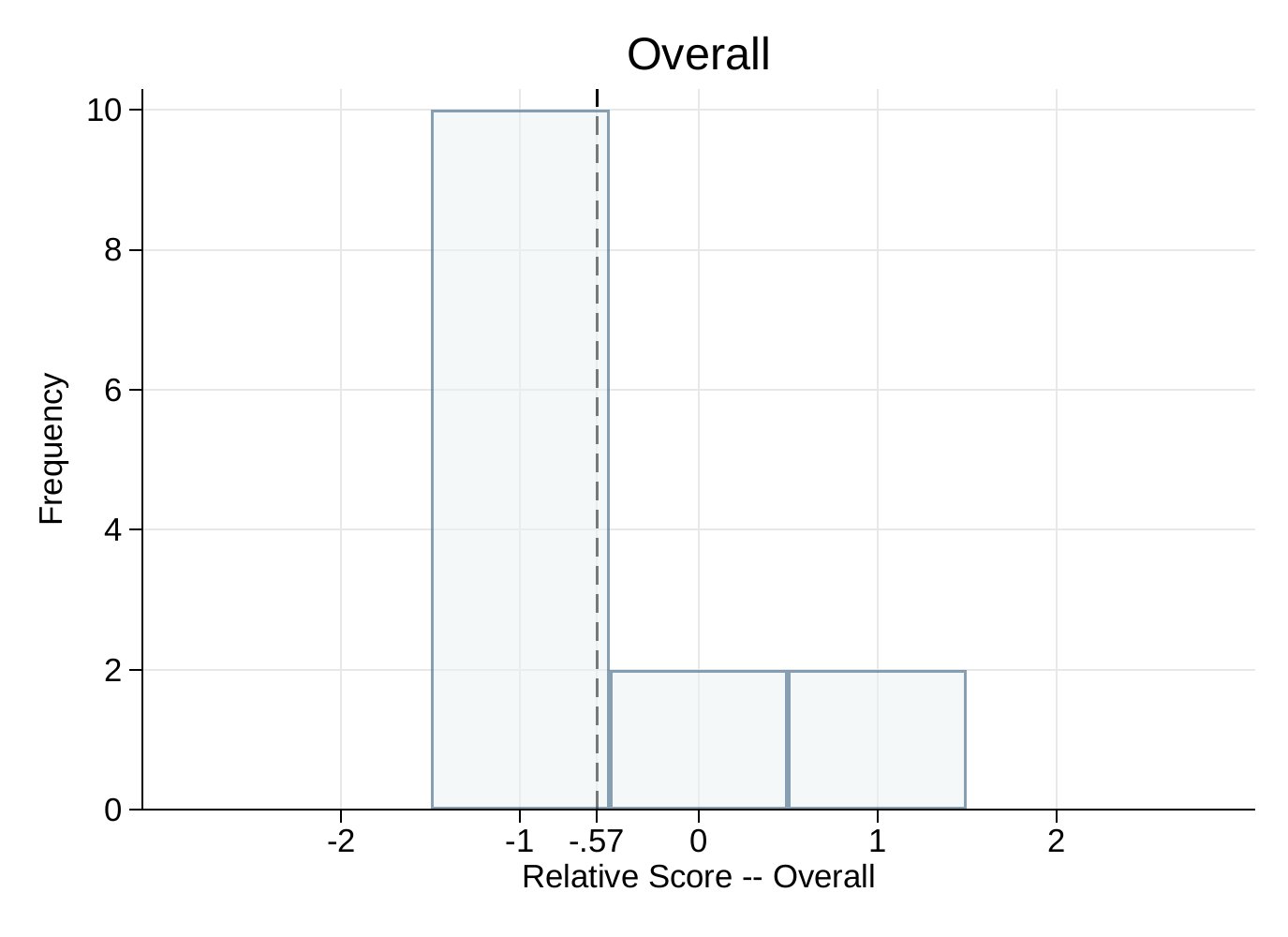}
    \label{fig:valid_main}

  \flushleft  {\footnotesize  \textit{Notes.} Histogram of relative summary quality scores from the blind human evaluation. Horizontal axis is the relative quality of the machine summaries, from low to high.}
\end{figure}

Qualitatively, the annotator could not easily tell which summaries were machine-generated, and felt both summaries were satisfactory in almost all cases. That is reflected in the relative scores, reported in Figure \ref{fig:valid_main}. Out of the fourteen cases, the ground-truth summary was slightly better for ten cases, the machine summary was slightly better for two cases, and they were subjectively equal for two cases. The mean relative score for the machine summary is -0.57. See Appendix \ref{app:qual} for further details.

Note that these prominent Supreme Court Cases are not a representative sample. They are relatively lengthy and complex, dealing with unusual issues and making novel holdings. Thus, this is one of the most difficult set of cases for the machine summarizer to deal with. Yet still, the resulting summaries are of impressive quality and almost as good as the human-generated summaries according to our trained lawyer.  We hope that more comprehensive assessments of subjective summary quality are produced in the future. 

\section{Conclusion}

In this paper, we tackled the extractive summarization of U.S. Supreme Court opinions. The main challenge is that such opinions are very long, which poses unique challenges for NLP methods. We make use of recent advances in extractive long-document summarization, and experiment with two different model families. We find that that the reinforcement-based MemSum model performs best on this task. We manually confirm the quality of these summaries on 14 landmark U.S. cases. We open-source our models which will help democratize legal practice by making legal primary sources more accessible to a broad audience.

\clearpage

\section*{Limitations}

\paragraph{Extractive Summarization.} 
Extractive summarization of long documents has known limitations, e.g., such summaries tend to be lengthy, incoherent, disconnected and can be verbose. While abstractive summarization does not suffer from these limitations, these models are known to hallucinate and produce non-factual content -- which is absolutely not acceptable in the case of summarizing court opinions. In a perfect world, we would tackle summarizing court opinions using faithful abstractive summarizations which we can guarantee do not hallucinate. Howver, the current technology does not support this, hence our decision to approach summarization in an extractive matter.

\paragraph{Limitations of Models.} 
While we believe that our summaries are of high quality, in our blind human evaluation, in 70\% the annotator preferred the human summary compared to our machine-generated summary -- indicating that there is still room for improvement. Summarizing long legal documents poses many challenges and the current state of NLP is not ready yet to tackle all of them. Given the tremendous amount of progress in summarizing long documents, we expect to re-visit summarizing U.S. court opinions in the future, using more advanced technology. 

\paragraph{Limitations of our Dataset.} 
Due to licensing issues, we cannot release the dataset we trained our model on. Also, we are only concerned with summarizing U.S. court opinions. In future work, we want to explore how much our models generalize to e.g., other jurisdictions and other types of legal documents. We have already gotten promising results on summarizing legislative documents.

\paragraph{Computational Budget.} 
Given our computational budget and the time it takes to fine-tune models on 400K long court opinions, we were not able to conduct more experiments than the ones presented in the paper. Given infinite time and compute budget, we would have (1) replaced MemSum's RNN backbone with a transformer architecture (2) explored other long-document transformer models and (3) would have experimented with various hyper-parameters in more detail.

\section*{Ethics Statement}

\paragraph{License and Intended Use.} 

Our code and trained models are available as open source software at \url{https://github.com/bauerem/legal_memsum}. We believe our models are a step towards democratizing U.S. law, thus our decision to release them for non-commercial use. The intended use of our models is legal research and legal journalism. Further, we allow the general public access to key information in U.S. court opinions and speed up the work of legal practitioners. The training data cannot be released due to license restrictions, but the trained models are released for public/social good.

\paragraph{Misuse Potential.}

\paragraph{Risks of this Work.} 
While we believe our models are useful, they will still produce machine-generated text. There is the risk that our models miss key sentences in the opinion, thus we advice practicioners using our models to always consult the original opinion as well. Furthermore, although our models are of extractive nature and thus by constructing cannot hallucinate, it is conceivable that the key findings generated by our summaries are presented out of context and thus might be misleading. Even though we have not found this in our human evaluation study, we do not exclude the risk of this happening at scale. 

\paragraph{Model Bias.} 
It is widely known that ML models suffer from picking up spurious correlations from data. Furthermore, it has been shown that language models (e.g., the word embeddings we use for MemSum and the LED encoder for our extractive model) suffer from inherent biases present in the pre-training data leading to biased models -- and we believe our models presented in this work also suffer from these biases.

\paragraph{Data Privacy.} 
The data used in this work are publicly available U.S. court opinions. There is no user-related data or private data involved.

\clearpage
\newpage

\bibliography{arxiv_bib}
\bibliographystyle{acl_natbib}

\clearpage

\appendix

\renewcommand{\thefigure}{A.\arabic{figure}} 
\setcounter{table}{0} \renewcommand{\thetable}{A.\arabic{table}} 

\begin{center}
    ONLINE APPENDIX
\end{center}

\section{Dataset Statistics}

\begin{table}[ht!]
    \centering
    \scriptsize
    \begin{tabular}{l c c c}
      & Average & Median & 90\%-quantile \\ \hline
      Words in Opinion & 2'745 & 2'016 & 5'440 \\ 
      Sentences in Opinion & 86 & 62 & 173 \\
      Words in Summary & 435 & 293 & 965 \\ 
      Sentences in Summary & 6 & 4 & 12 \\ \hline
    \end{tabular}
    \caption{Dataset Statistics}
    \label{tab:dataset_stats}
\end{table}

\section{MemSum Details} \label{app:memsum}

\paragraph{Policy.}

The policy, i.e., the probability distributions of the actions in MemSum is computed according to Equation \eqref{eq:policy}.

\begin{equation}
    \label{eq:policy}
    \resizebox{\linewidth}{!}{
    \begin{math}
    \begin{aligned}
    \pi(A_t|S_t,\theta_t) & = p(\texttt{stop}|S_t,\theta_t)p(a_t|\texttt{stop},S_t,\theta_t) \\
    p(a_t|\texttt{stop},S_t,\theta_t) & = 
    \begin{cases} 
        \frac{u_{a_t}(S_t, \theta_t)}{\sum_{j\in I_t} u_j(S_t, \theta_t)} 
        & \text{if $\texttt{stop}=\texttt{false}$} \\
        \frac{1}{|I_t|}
        & \text{if $\texttt{stop}=\texttt{true}$}
        \end{cases}.
    \end{aligned}
    \end{math}
    }
\end{equation}

The REINFORCE algorithm \cite{REINFORCE, Sutton1998} maximizes the value function $J(\theta)\propto \E_{\pi_\theta}[G_t\nabla \log \pi_\theta]$, where $R_k$ is the Markov Decision Process' reward at the $k$-th i, $\gamma$ is the reward discount rate and $G_t=\sum^T_{k=t+1}\gamma^{k-t-1}R_k$ is the advantage function of discounted future rewards. This is achieved \cite{Sutton1998} by way of the optimization scheme

\begin{equation*}
    \theta\leftarrow \theta + \alpha \gamma^t G_t \nabla \log \pi_\theta.
\end{equation*}

In the case of MemSum, $R_t=0$ for all $t<T$and $
    R_T=
        \frac{1}{3}(\text{ROUGE-1}_f + \text{ROUGE-2}_f + \text{ROUGE-L}_f)
$. In addition, a factor of $1/(T+1)$ is introduced into the increment to encourage compact summaries \cite{luo-etal-2019-reading} and the discount factor $\gamma$ was set to $1$. so $\gamma^t G_t=\frac{R_T}{T+1}$. Therefore, we obtain the parameter update rule:

\begin{equation}
\label{eq:memsum_parameter_updating_rule}
    \theta\leftarrow \theta + \alpha \frac{R_T}{T+1} \nabla \log \pi_\theta.
\end{equation}

\paragraph{Training.}

The policy network was trained according to the policy gradient method described in \citet{Sutton1998}. Following \citet{narayan-etal-2018-ranking, mohsen-2020-hsn} We sampled an episode $(S_0,A_0,R_1,\dots,S_T,R_T)$ with a high ROUGE score. At each step $t$, we encode the state information $S_t$ using the local sentence encoder, the global context encoder and the extraction history encoder of MemSum and compute the probability of the current action based on the policy $\pi(A_t|S_t,\theta)$. After computing the action probability for all time steps in an episode, we update the policy parameters according to Equation \eqref{eq:memsum_parameter_updating_rule}. The training algorithm of MemSum was described in detail in the original MemSum paper \citet{MemSum} and we refer the interested reader to this paper for the full algorithm. 

\paragraph{Stopping Probability Hyperparameter.}

\begin{figure}[H]
    \centering
    \includegraphics[width=7cm]{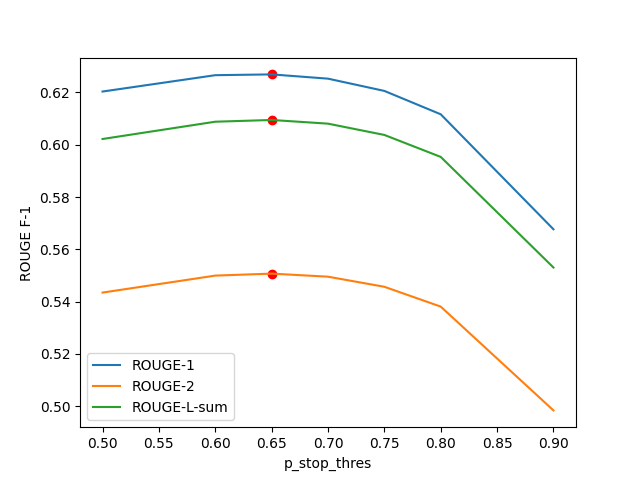}
    \caption{Rouge Score on (measured on validation set) for different stopping thresholds.}
    \label{fig:stopping}
\end{figure}

We tune the value of the stopping probability threshold hyperparameter using the training/validation set. We find that ROUGE scores are somewhat robust to the choice of this hyperparameter (as long as it is below $p<0.8$), see Figure . From this figure, we observe that highest ROUGE scores are obtained for selecting $p=0.65$, hence we implement our algorithm with this value. 

\begin{figure}[H]
    \centering
    \includegraphics[width=7cm]{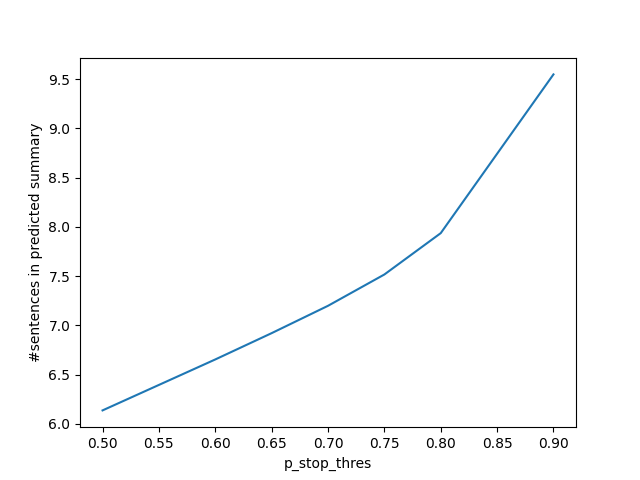}
    \caption{Number of sentences extracted per opinion for different stopping thresholds. Using the threshold yielding the highest ROUGE scores (0.65) yields on average 7 sentences per summary.}
    \label{fig:stopsent}
\end{figure}

We also plot the amount of sentences extracted for different stopping thresholds, and if we set the threshold to $p=0.5$, we extract on average 6 sentences. See Figure \ref{fig:stopsent}. Setting the threshold to $p=0.65$ yields 6.8 sentences on average, resulting in a compression ratio of 14.7\%. This compression ratio is close to the compression ratio of 15.8\% of the ground-truth summaries. 

\section{LawFormer Details} \label{app:lawformer}

For the ``LawFormer'' variant of the transformer summarizer, we further pre-train a BART checkpoint on 6Mio US court opinions. We crawled 5Mio district court opinions and 1Mio circuit court opinions, equalling 6.5B tokens or 35GB of text. We randomly mask 30\% of the tokens and randomly permute 30\% of the sentences in these opinions and our pre-training objective is to recover the original sequence from the corrupted input. The model is trained for one epoch on 4 GPUs in parallel using the transformers library \cite{wolf-etal-2020-transformers} and deepspeed \cite{deepspeed}, taking 7 days. This pre-trained model is then further fine-tuned on the summarization task.

\section{Greedy Oracle Extraction Algorithm} \label{sec:oracle_details}

Our gold summary labels are provided as plain text rather than a list of sentence identifiers. Small formatting/style differences precludes an exact match to identify the set of reference sentences in the original documents. Doing fuzzy pairwise text matches of all summary sentences and all full-document sentences is too computationally expensive. Hence, we used the greedy oracle algorithm \cite{MemSum} to approximate the upper limit of an oracle extractive summarizer. The greedy oracle algorithm works as follows. 

Given a document $\mathcal{D}$ with $N$ sentences $\{s_i|i=1,\dots,N\}$ and a ground-truth summary $\mathcal{A}_g$, an ideal oracle extractive summarizer selects $M$ ($M$ is a varying number, and typically $M\ll N$) sentences out of the $N$ sentences in $\mathcal{D}$ to construct an oracle extractive summary $\mathcal{A}_o$, such that the ROUGE score (e.g., average of the ROUGE-1 and ROUGE-2 F1 scores) $R(\mathcal{A}_g, \mathcal{A}_o)$ is maximal. Mathematically, this can be described as:
\begin{equation}
    \begin{aligned}
    \mathcal{A}_o = \arg\max_{\mathcal{A} \subseteq \mathcal{D}, |\mathcal{A}| = M, |\mathcal{D}| = N} R(\mathcal{A}_g, \mathcal{A})
    \end{aligned}
\end{equation}

A straightforward method for determining $\mathcal{A}_o$ involves enumerating all possible combinations of $M$ sentences from the set of $N$ sentences comprising $\mathcal{D}$, resulting in ${N \choose M}$ candidate extractive summaries. 
The ROUGE score is subsequently computed for each candidate summary, and the summary with the highest score is designated as the oracle summary. However, this approach is computationally intensive, owing to the large number of combinations represented by ${N \choose M}$. As such, the need for a more efficient approximation algorithm is apparent.

The greedy oracle algorithm approximates the real oracle in the following way. When summarizing the document $\mathcal{D}$, the greedy oracle selects sentences from $\mathcal{D}$ iteratively in a greedy manner. At the initial step, when no sentence has been selected and $\mathcal{A}_o$ is an empty set, the greedy oracle selects from $\mathcal{D}$ a sentence $s_i$ whose ROUGE score with respect to the ground-truth summary $\mathcal{A}_g$ is maximal among all sentences. In the next step, the greedy oracle selects a sentence $s_j$ from the remaining sentences in $\mathcal{D}$ that satisfies the following criterion: Compared with any other remaining sentence $s_{k (k\neq j)}$, after adding $s_j$ to $\mathcal{A}_o$, the ROUGE score $R(\mathcal{A}_g, \mathcal{A}_o)$ is maximally increased. The greedy oracle continues with this process until reaching a certain step when adding any more sentences to $\mathcal{A}_o$ will result in a drop in $R(\mathcal{A}_g, \mathcal{A}_o)$, then the extraction is stopped and the current $\mathcal{A}_o$ is treated as an approximation of the real oracle summary. This process is summarized in Algorithm \ref{alg:greedy_oracle}.

\begin{algorithm}
    \textbf{Input:} $\mathcal{D}$,$\mathcal{A}_g$\\
    \textbf{Initialize:} $\mathcal{A}_o \leftarrow \emptyset$, $MaxStep \leftarrow |\mathcal{D}|$
  \begin{algorithmic}[1]
    \For{$step$ in $1,2,\dots,MaxStep$}
        \If {$ \forall s_k \in \mathcal{D}, R(\mathcal{A}_g, \mathcal{A}_o \cup \{s_k\}) - R(\mathcal{A}_g, \mathcal{A}_o)\leq 0$ }
            \State \textbf{break}
        \EndIf
    
        \State $s_j\leftarrow \arg\max_{s_k\in \mathcal{D}} R(\mathcal{A}_g, \mathcal{A}_o \cup \{s_k\}) - R(\mathcal{A}_g, \mathcal{A}_o)$
        \State $\mathcal{A}_o \leftarrow \mathcal{A}_o \cup \{s_j\}$
        \State $\mathcal{D} \leftarrow \mathcal{D} \setminus \{s_j\}$
    \EndFor
    \\
 \textbf{return} $\mathcal{A}_o$
  \end{algorithmic} 
  \caption{Greedy Oracle Algorithm}
  \label{alg:greedy_oracle}
\end{algorithm}

\section{Additional Material for Qualitative Validation} \label{app:qual}

\begin{figure}
    \centering
    
    \includegraphics[width=\columnwidth]{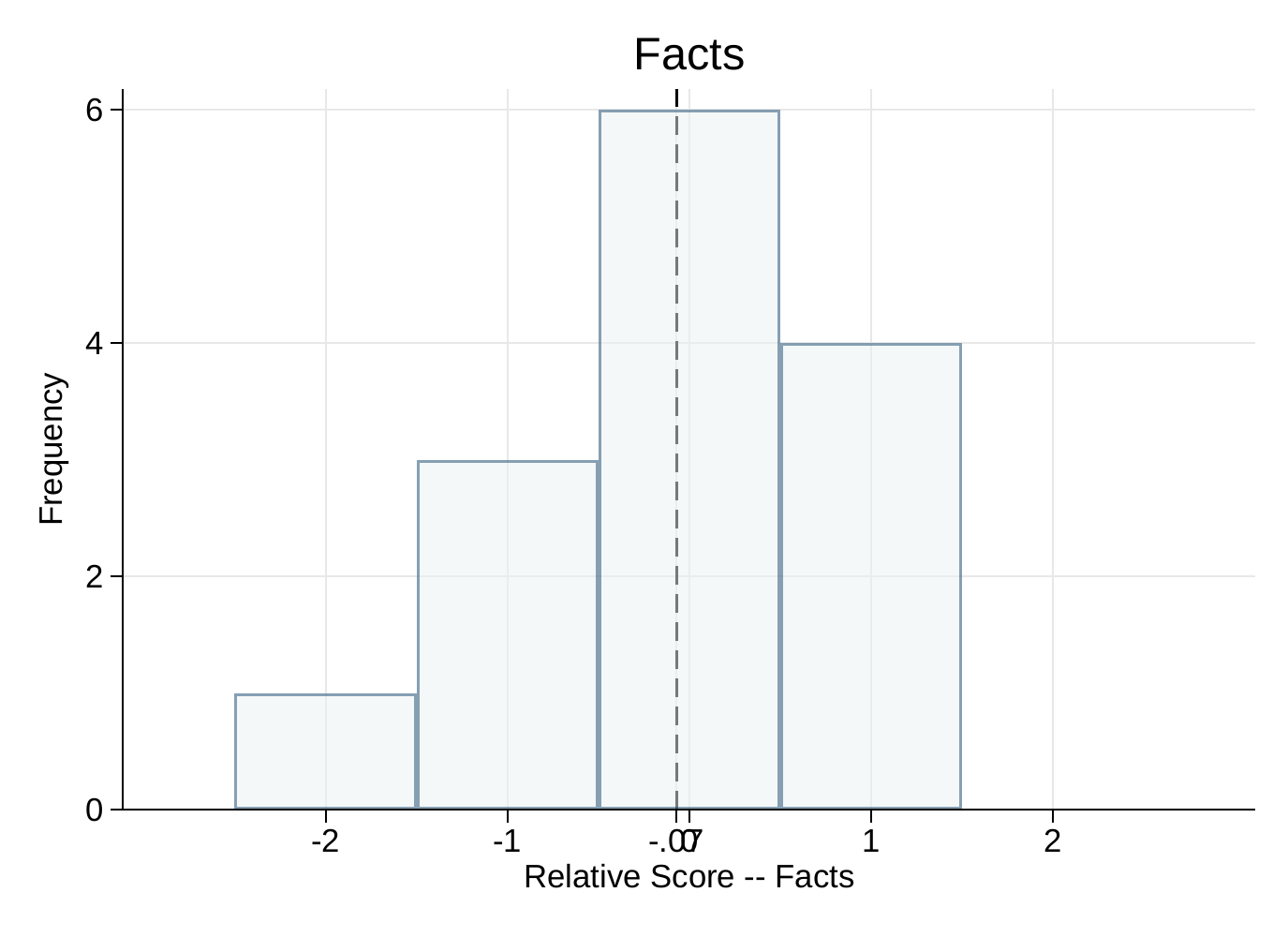}
    
    \includegraphics[width=\columnwidth]{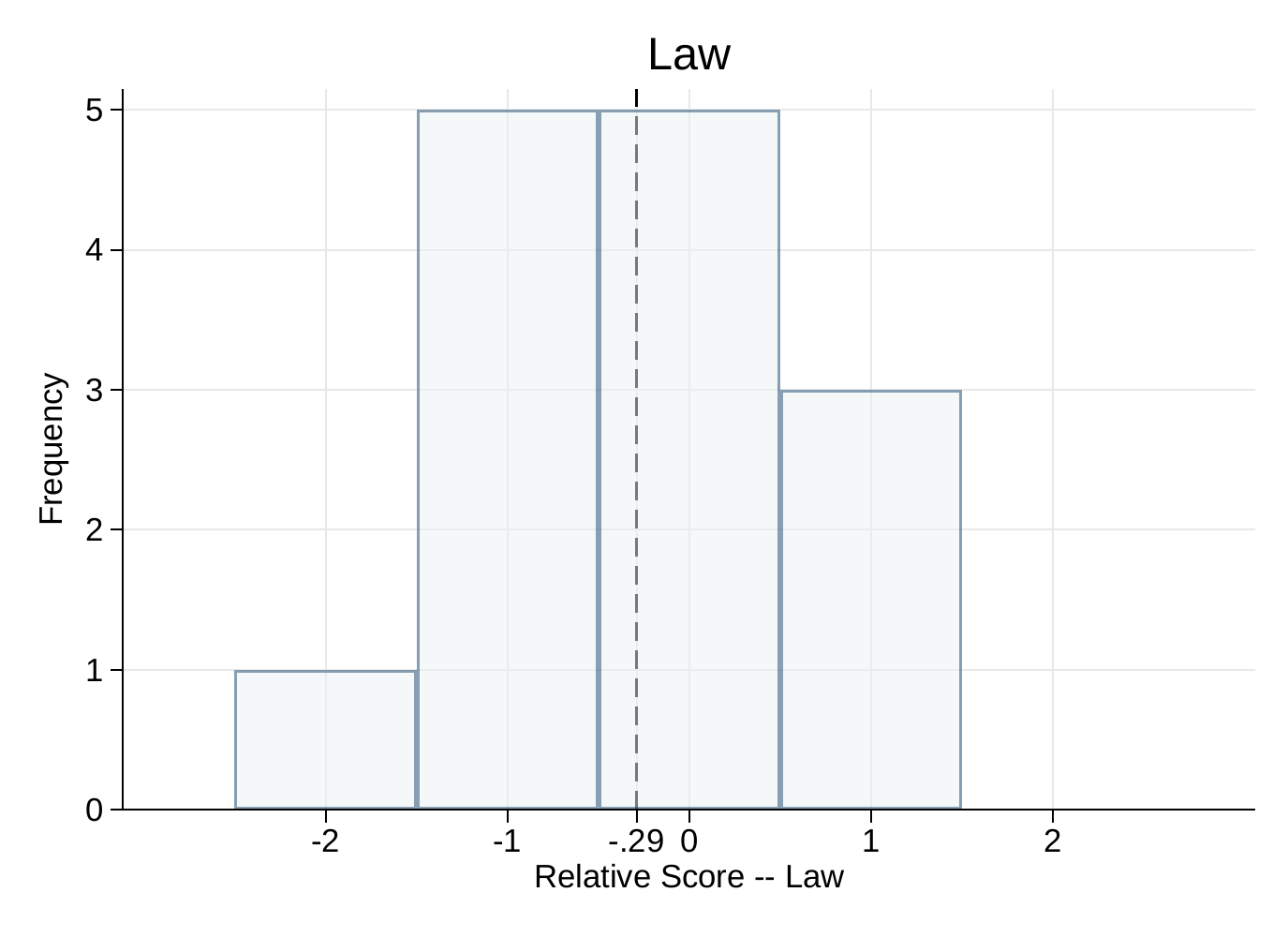}

    \includegraphics[width=\columnwidth]{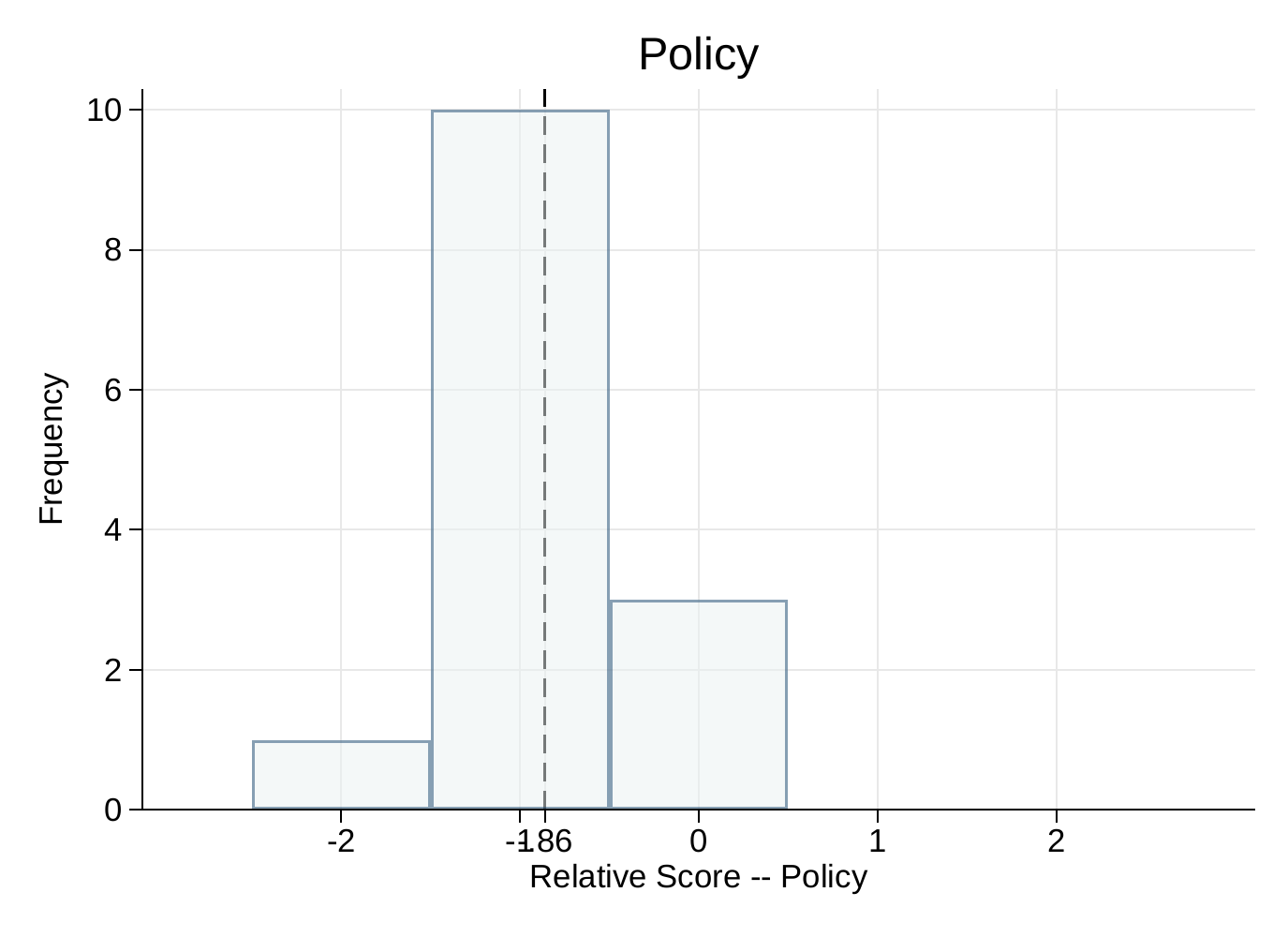}
    
    \includegraphics[width=\columnwidth]{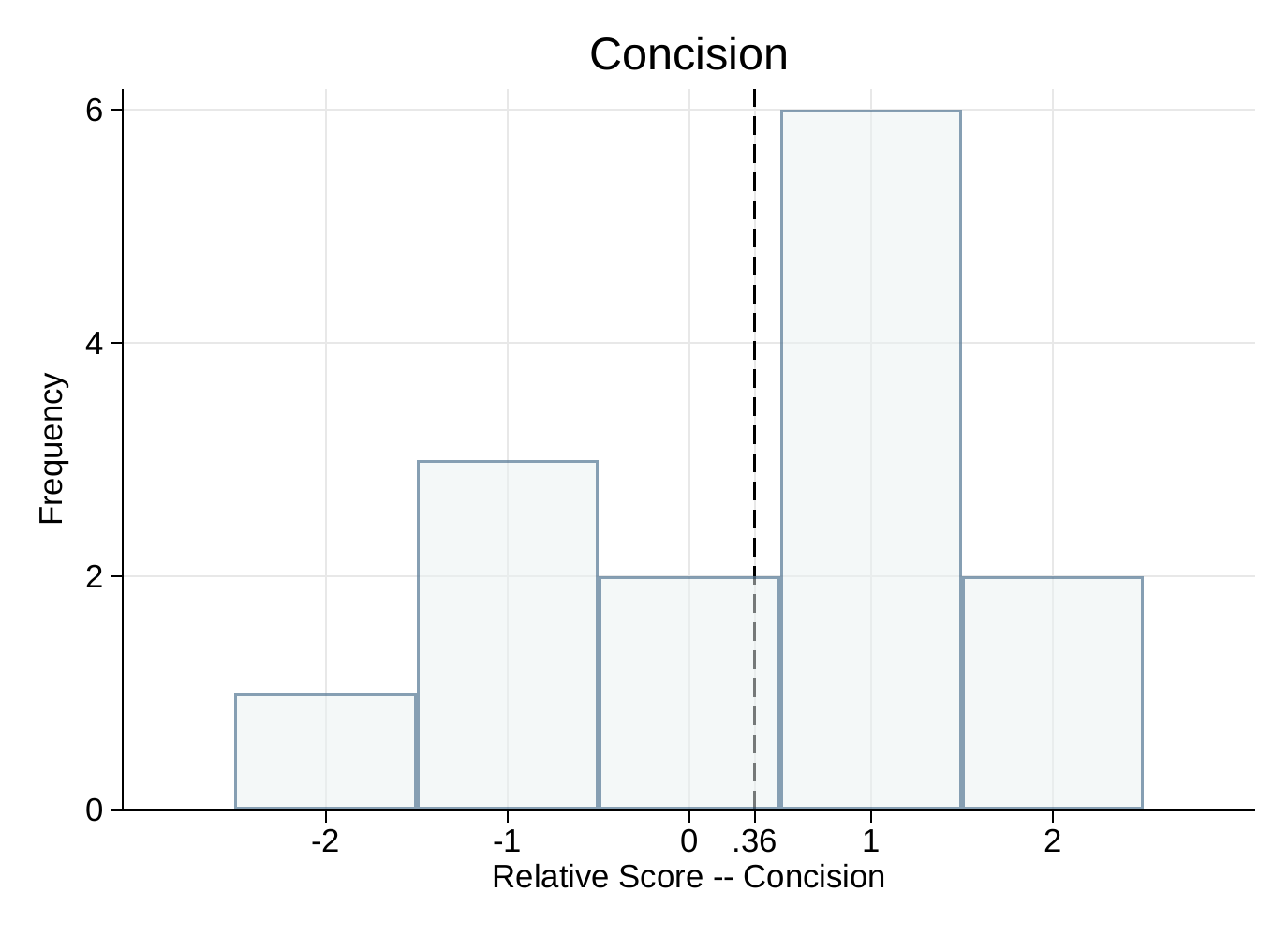}
    
    \caption{Human Validation: Other Dimensions of Quality} \label{fig:valid_app}
\end{figure}

The annotator scored the summaries along five dimensions of summary quality: facts, law, policy, concision, and overall. Figure \ref{fig:valid_app} shows the validation results for the other dimensions of summary quality: in order, the summary of the facts, law, and policy features of the opinion, and finally the concision (shortness) of the summary. We see that human and machine summaries were equally good at summarizing case facts. The human summaries were better at summarizing the law and policy dimensions of the cases. The machine summaries tended to be more concise. 

Finally, Table \ref{tab:summaries_Brown_v._Bd._of_Educ.} shows machine-produced extractive summaries for the majority opinions from a selection of 14 prominent U.S. Supreme Court cases. These cases come from a list of 25 cases that have been cited most often by U.S. Law Review articles from \href{https://home.heinonline.org/blog/2018/09/most-cited-u-s-supreme-court-cases-in-heinonline-part-iii/}{HeinOnline}. In the first column, we list the rank of the extraction order as a proxy for sentence importance (the earlier a sentence is extracted in MemSum, the higher it's importance). In the second column, we list the sentence position in the original case. In the third column, we show the actual sentence.

\clearpage
\newpage

\begin{table*}[]
    \centering
    \caption{Extracted Sentences from Prominent U.S. Supreme Court Opinions} \label{tab:summaries_Brown_v._Bd._of_Educ.}
    \tiny
    \begin{tabular}{c c | p{14.0cm}}
    \multicolumn{3}{c}{\textbf{Brown v. Bd. of Educ. (156 sentences in original)}} \\
Rank & Position & Extracted sentence \\ \hline
5 & 47 & Under that doctrine, equality of treatment is accorded when the races are provided substantially equal facilities, even though these facilities be separate.\\
3 & 83 & 5Slaughter-House Cases, 16 Wall. 36, 67-72 (1873); Strauder v. West Virginia, 100 U.S. 303, 307-308 (1880):"It ordains that no State shall deprive any person of life, liberty, or property, without due process of law, or deny to any person within its jurisdiction the equal protection of the laws.\\
2 & 84 & What is this but declaring that the law in the States shall be the same for the black as for the white; that all persons, whether colored or white, shall stand equal before the laws of the States, and, in regard to the colored race, for whose protection the amendment was primarily designed, that no discrimination shall be made against them by law because of their color?\\
1 & 85 & The words of the amendment, it is true, are prohibitory, but they contain a necessary implication of a positive immunity, or right, most valuable to the colored race, -the right to exemption from unfriendly legislation against them distinctively as colored, -exemption from legal discriminations, implying inferiority in civil society, lessening the security of their enjoyment of the rights which others enjoy, and discriminations which are steps towards reducing them to the condition of a subject race."See also Virginia v. Rives, 100 U.S. 313, 318 (1880); Ex parte Virginia, 100 U.S. 339, 344-345 (1880).\\
8 & 119 & Today it is a principal instrument in awakening the child to cultural values, in preparing him for later professional training, and in helping him to adjust normally to his environment.\\
7 & 120 & In these days, it is doubtful that any child may reasonably be expected to succeed in life if he is denied the opportunity of an education.\\
6 & 121 & Such an opportunity, where the state has undertaken to provide it, is a right which must be made available to all on equal terms.\\
9 & 122 & We come then to the question presented: Does segregation of children in public schools solely on the basis of race, even though the physical facilities and other "tangible" factors may be equal, deprive the children of the minority group of equal educational opportunities?\\
10 & 132 & Segregation with the sanction of law, therefore, has a tendency to [retard] the educational and mental development of negro children and to deprive them of some of the benefits they would receive in a racial[ly] integrated school system."\\
4 & 150 & Assuming it is decided that segregation in public schools violates the Fourteenth Amendment"(a) would a decree necessarily follow providing that, within the limits set by normal geographic school districting, Negro children should forthwith be admitted to schools of their choice, or"(b) may this Court, in the exercise of its equity powers, permit an effective gradual adjustment to be brought about from existing segregated systems to a system not based on color distinctions?"5.\\ \\
\end{tabular}
   
    \begin{tabular}{c c | p{14.0cm}}
    \multicolumn{3}{c}{\textbf{Roe v. Wade (619 sentences in original)}} \\
Rank & Position & Extracted sentence \\ \hline
10 & 5 & One's philosophy, one's experiences, one's exposure to the raw edges of human existence, one's religious training, one's attitudes toward life and family and their values, and the moral standards one establishes and seeks to observe, are all likely to influence and to color one's thinking and conclusions about abortion.\\
9 & 7 & Our task, of course, is to resolve the issue by constitutional measurement, free of emotion and of predilection.\\
8 & 71 & We conclude, nevertheless, that those decisions do not foreclose our review of both the injunctive and the declaratory aspects of a case of this kind when it is properly here, as this one is, on appeal under § 1253 from specific denial of injunctive relief, and the arguments as to both aspects are necessarily identical.\\
1 & 91 & The usual rule in federal cases is that an actual controversy must exist at stages of appellate or certiorari review, and not simply at the date the action is initiated.\\
3 & 92 & United States v. Munsingwear, Inc., 340 U.S. 36 (1950); Golden v. Zwickler, supra;SEC v. Medical Committee for Human Rights, 404 U.S. 403 (1972).But when, as here, pregnancy is a significant fact in the litigation, the normal 266-day human gestation period is so short that the pregnancy will come to term before the usual appellate process is complete.\\
2 & 93 & If that termination makes a case moot, pregnancy litigation seldom will survive much beyond the trial stage, and appellate review will be effectively denied.\\
5 & 94 & Our law should not be that rigid.\\
4 & 95 & Pregnancy often comes more than once to the same woman, and in the general population, if man is to survive, it will always be with us.\\
6 & 96 & Pregnancy provides a classic justification for a conclusion of nonmootness.\\
7 & 97 & It truly could be "capable of repetition, yet evading review."\\ \\
\end{tabular}
    
    \begin{tabular}{c c | p{14.0cm}}
    \multicolumn{3}{c}{\textbf{Erie R.R. v. Tompkins (286 sentences in original)}} \\
Rank & Position & Extracted sentence \\ \hline
6 & 17 & Where the public has made open and notorious use of a railroad right of way for a long period of time and without objection, the company owes to persons on such permissive pathway a duty of care in the operation of its trains.\\
2 & 19 & It is likewise generally recognized law that a jury may find that negligence exists toward a pedestrian using a permissive path on the railroad right of way if he is hit by some object projecting from the side of the train."\\
5 & 22 & First. Swift v. Tyson, 16 Pet. 1, 18, held that federal courts exercising jurisdiction on the ground of diversity of citizenship need not, in matters of general jurisprudence, apply the unwritten law of the State as declared by its highest court; that they are free to exercise an independent judgment as to what the common law of the State is -or should be; and that, as there stated by Mr. Justice Story:"the true interpretation of the thirty-fourth section limited its application to state laws strictly local, that is to say, to the positive statutes of the state, and the construction thereof adopted by the local tribunals, and to rights and titles to things having a permanent locality, such as the rights and titles to real estate, and other matters immovable and intraterritorial in their nature and character.\\
3 & 23 & It never has been supposed by us, that the section did apply, or was intended to apply, to questions of a more general nature, not at all dependent upon local statutes or local usages of a fixed and permanent operation, as, for example, to the construction of ordinary contracts or other written instruments, and especially to questions of general commercial law, where the state tribunals are called upon to perform the like functions as ourselves, that is, to ascertain upon general reasoning and legal analogies, what is the true exposition of the contract or instrument, or what is the just rule furnished by the principles of commercial law to govern the case."\\
7 & 80 & the liability for torts committed within the State upon persons resident or property located there, even where the question of liability depended upon the scope of a property right conferred by the State; 13Chicago v. Robbins, 2 Black 418, 428.\\
1 & 124 & The doctrine rests upon the assumption that there is "a transcendental body of law outside of any particular State but obligatory within it unless and until changed by statute," that federal courts have the power to use their judgment as to what the rules of common law are; and that in the federal courts "the parties are entitled to an independent judgment on matters of general law":"but law in the sense in which courts speak of it today does not exist without some definite authority behind it.\\
9 & 125 & The common law so far as it is enforced in a State, whether called common law or not, is not the common law generally but the law of that State existing by the authority of that State without regard to what it may have been in England or anywhere else.\\
8 & 172 & In all the various cases, which have hitherto come before us for decision, this Court have uniformly supposed, that the true interpretation of the thirty-fourth section limited its application to state laws strictly local, that is to say, to the positive statutes of the state, and the construction thereof adopted by the local tribunals, and to rights and titles to things having a permanent locality, such as the rights and titles to real estate, and other matters immovable and intraterritorial in their nature and character.\\
4 & 173 & It never has been supposed by us, that the section did apply, or was designed to apply, to questions of a more general nature, not at all dependent upon local statutes or local usages of a fixed and permanent operation, as, for example, to the construction of ordinary contracts or other written instruments, and especially to questions of general commercial law, where the state tribunals are called upon to perform the like functions as ourselves, that is, to ascertain upon general reasoning and legal analogies, what is the true exposition of the contract or instrument, or what is the just rule furnished by the principles of commercial law to govern the case.\\
10 & 174 & And we have not now the slightest difficulty in holding, that this section, upon its true intendment and construction, is strictly limited to local statutes and local usages of the character before stated, and does not extend to contracts and other instruments of a commercial nature, the true interpretation and effect whereof are to be sought, not in the decisions of the local tribunals, but in the general principles and doctrines of commercial jurisprudence.\\ \\
\end{tabular}

    \begin{tabular}{c c | p{14.0cm}}
    \multicolumn{3}{c}{\textbf{Chevron, U.S.A., Inc. v. NRDC, Inc. (341 sentences in original)}} \\
Rank & Position & Extracted sentence \\ \hline
6 & 44 & If the intent of Congress is clear, that is the end of the matter; for the court, as well as the agency, must give effect to the unambiguously expressed intent of Congress.\\
7 & 47 & If a court, employing traditional tools of statutory construction, ascertains that Congress had an intention on the precise question at issue, that intention is the law and must be given effect.\\
3 & 48 & If, however, the court determines Congress has not directly addressed the precise question at issue, the court does not simply impose its own construction on the statute, 10See generally, R. Pound, The Spirit of the Common Law 174-175 (1921). as would be necessary in the absence of an administrative interpretation.\\
5 & 49 & Rather, if the statute is silent or ambiguous with respect to the specific issue, the question for the court is whether the agency's answer is based on a permissible construction of the statute.\\
4 & 50 & 11The court need not conclude that the agency construction was the only one it permissibly could have adopted to uphold the construction, or even the reading the court would have reached if the question initially had arisen in a judicial proceeding.\\
10 & 52 & "The power of an administrative agency to administer a congressionally created... program necessarily requires the formulation of policy and the making of rules to fill any gap left, implicitly or explicitly, by Congress."\\
9 & 54 & If Congress has explicitly left a gap for the agency to fill, there is an express delegation of authority to the agency to elucidate a specific provision of the statute by regulation.\\
8 & 57 & In such a case, a court may not substitute its own construction of a statutory provision for a reasonable interpretation made by the administrator of an agency.\\
1 & 60 & and the principle of deference to administrative interpretations"has been consistently followed by this Court whenever decision as to the meaning or reach of a statute has involved reconciling conflicting policies, and a full understanding of the force of the statutory policy in the given situation has depended upon more than ordinary knowledge respecting the matters subjected to agency regulations.\\
2 & 62 & If this choice represents a reasonable accommodation of conflicting policies that were committed to the agency's care by the statute, we should not disturb it unless it appears from the statute or its legislative history that the accommodation is not one that Congress would have sanctioned."\\ \\
\end{tabular}

\end{table*}

\begin{table*}
    \tiny 
    \begin{tabular}{c c | p{14.0cm}}
    \multicolumn{3}{c}{\textbf{Lochner v. New York (138 sentences in original)}} \\
Rank & Position & Extracted sentence \\ \hline
5 & 6 & The mandate of the statute that "no employe shall be required or permitted to work," is the substantial equivalent of an enactment that "no employe shall contract or agree to work," more than ten hours per day, and as there is no provision for special emergencies the statute is mandatory in all cases.\\
2 & 10 & The general right to make a contract in relation to his business is part of the liberty of the individual protected by the Fourteenth Amendment of the Federal Constitution.\\
6 & 12 & Under that provision no State can deprive any person of life, liberty or property without due process of law.\\
8 & 13 & The right to purchase or to sell labor is part of the liberty protected by this amendment, unless there are circumstances which exclude the right.\\
10 & 16 & Both property and liberty are held on such reasonable conditions as may be imposed by the governing power of the State in the exercise of those powers, and with such conditions the Fourteenth Amendment was not designed to interfere.\\
9 & 18 & The State, therefore, has power to prevent the individual from making certain kinds of contracts, and in regard to them the Federal Constitution offers no protection.\\
7 & 19 & If the contract be one which the State, in the legitimate exercise of its police power, has the right to prohibit, it is not prevented from prohibiting it by the Fourteenth Amendment.\\
3 & 20 & Contracts in violation of a statute, either of the Federal or state government, or a contract to let one's property for immoral purposes, or to do any other unlawful act, could obtain no protection from the Federal Constitution, as coming under the liberty of person or of free contract.\\
4 & 21 & Therefore, when the State, by its legislature, in the assumed exercise of its police powers, has passed an act which seriously limits the right to labor or the right of contract in regard to their means of livelihood between persons who are sui juris (both employer and employe), it becomes of great importance to determine which shall prevail -the right of the individual to labor for such time as he may choose, or the right of the State to prevent the individual from laboring or from entering into any contract to labor, beyond a certain time prescribed by the State.\\
1 & 129 & The purpose of a statute must be determined from the natural and legal effect of the language employed; and whether it is or is not repugnant to the Constitution of the United States must be determined from the natural effect of such statutes when put into operation, and not from their proclaimed purpose.\\ \\
\end{tabular}

    \begin{tabular}{c c | p{14.0cm}}
    \multicolumn{3}{c}{\textbf{Katz v. United States (144 sentences in original)}} \\
Rank & Position & Extracted sentence \\ \hline
1 & 4 & Whoever being engaged in the business of betting or wagering knowingly uses a wire communication facility for the transmission in interstate or foreign commerce of bets or wagers or information assisting in the placing of bets or wagers on any sporting event or contest, or for the transmission of a wire communication which entitles the recipient to receive money or credit as a result of bets or wagers, or for information assisting in the placing of bets or wagers, shall be fined not more than \$ 10,000 or imprisoned not more than two years, or both."(b)\\
2 & 5 & Nothing in this section shall be construed to prevent the transmission in interstate or foreign commerce of information for use in news reporting of sporting events or contests, or for the transmission of information assisting in the placing of bets or wagers on a sporting event or contest from a State where betting on that sporting event or contest is legal into a State in which such betting is legal."\\
8 & 27 & Whether a public telephone booth is a constitutionally protected area so that evidence obtained by attaching an electronic listening recording device to the top of such a booth is obtained in violation of the right to privacy of the user of the booth.\\
7 & 35 & ... And a person can be just as much, if not more, irritated, annoyed and injured by an unceremonious public arrest by a policeman as he is by a seizure in the privacy of his office or home."\\
10 & 44 & But the protection of a person's general right to privacy -his right to be let alone by other people 6See Warren \& Brandeis, The Right to Privacy, 4 Harv. L. Rev. 193 (1890). -is, like the protection of his property and of his very life, left largely to the law of the individual States.\\
6 & 57 & What a person knowingly exposes to the public, even in his own home or office, is not a subject of Fourth Amendment protection.\\
4 & 73 & Indeed, we have expressly held that the Fourth Amendment governs not only the seizure of tangible items, but extends as well to the recording of oral statements, overheard without any "technical trespass under... local property law."\\
3 & 75 & Once this much is acknowledged, and once it is recognized that the Fourth Amendment protects people -and not simply "areas" -against unreasonable searches and seizures, it becomes clear that the reach of that Amendment cannot turn upon the presence or absence of a physical intrusion into any given enclosure.\\
9 & 101 & Rule 41 (d) does require federal officers to serve upon the person searched a copy of the warrant and a receipt describing the material obtained, but it does not invariably require that this be done before the search takes place.\\
5 & 114 & Searches conducted without warrants have been held unlawful "notwithstanding facts unquestionably showing probable cause," Agnello v. United States, 269 U.S. 20, 33, for the Constitution requires "that the deliberate, impartial judgment of a judicial officer... be interposed between the citizen and the police.\\ \\
\end{tabular}

    \begin{tabular}{c c | p{14.0cm}}
    \multicolumn{3}{c}{\textbf{Plessy v. Ferguson (95 sentences in original)}} \\
Rank & Position & Extracted sentence \\ \hline
3 & 3 & The first section of the statute enacts "that all railway companies carrying passengers in their coaches in this State, shall provide equal but separate accommodations for the white, and colored races, by providing two or more passenger coaches for each passenger train, or by dividing the passenger coaches by a partition so as to secure separate accommodations:\\
4 & 4 & Provided, That this section shall not be construed to apply to street railroads.\\
2 & 5 & No person or persons, shall be admitted to occupy seats in coaches, other than, the ones, assigned, to them on account of the race they belong to."\\
1 & 6 & By the second section it was enacted "that the officers of such passenger trains shall have power and are hereby required to assign each passenger to the coach or compartment used for the race to which such passenger belongs; any passenger insisting on going into a coach or compartment to which by race he does not belong, shall be liable to a fine of twenty-five dollars, or in lieu thereof to imprisonment for a period of not more than twenty days in the parish prison, and any officer of any railroad insisting on assigning a passenger to a coach or compartment other than the one set aside for the race to which said passenger belongs, shall be liable to a fine of twenty-five dollars, or in lieu thereof to imprisonment for a period of not more than twenty days in the parish prison; and should any passenger refuse to occupy the coach or compartment to which he or she is assigned by the officer of such railway, said officer shall have power to refuse to carry such passenger on his train, and for such refusal neither he nor the railway company which he represents shall be liable for damages in any of the courts of this State."\\
5 & 7 & The third section provides penalties for the refusal or neglect of the officers, directors, conductors and employes of railway companies to comply with the act, with a proviso that "nothing in this act shall be construed as applying to nurses attending children of the other race."\\
10 & 14 & Slavery implies involuntary servitude -a state of bondage; the ownership of mankind as a chattel, or at least the control of the labor and services of one man for the benefit of another, and the absence of a legal right to the disposal of his own person, property and services.\\
6 & 22 & By the Fourteenth Amendment, all persons born or naturalized in the United States, and subject to the jurisdiction thereof, are made citizens of the United States and of the State wherein they reside; and the States are forbidden from making or enforcing any law which shall abridge the privileges or immunities of citizens of the United States, or shall deprive any person of life, liberty or property without due process of law, or deny to any person within their jurisdiction the equal protection of the laws.\\
8 & 49 & Positive rights and privileges are undoubtedly secured by the Fourteenth Amendment; but they are secured by way of prohibition against state laws and state proceedings affecting those rights and privileges, and by power given to Congress to legislate for the purpose of carrying such prohibition into effect; and such legislation must necessarily be predicated upon such supposed state laws or state proceedings, and be directed to the correction of their operation and effect."\\
7 & 73 & While this was the case of a municipal ordinance, a like principle has been held to apply to acts of a state legislature passed in the exercise of the police power.\\
9 & 76 & In determining the question of reasonableness it is at liberty to act with reference to the established usages, customs and traditions of the people, and with a view to the promotion of their comfort, and the preservation of the public peace and good order.\\ \\
\end{tabular}

    \begin{tabular}{c c | p{14.0cm}}
    \multicolumn{3}{c}{\textbf{Loving v. Virginia (109 sentences in original)}} \\
Rank & Position & Extracted sentence \\ \hline
7 & 64 & The mere fact of equal application does not mean that our analysis of these statutes should follow the approach we have taken in cases involving no racial discrimination where the Equal Protection Clause has been arrayed against a statute discriminating between the kinds of advertising which may be displayed on trucks in New York City, Railway Express Agency, Inc. v. New York, 336 U.S. 106 (1949), or an exemption in Ohio's ad valorem tax for merchandise owned by a nonresident in a storage warehouse, Allied Stores of Ohio, Inc. v. Bowers, 358 U.S. 522 (1959).\\
10 & 84 & There can be no question but that Virginia's miscegenation statutes rest solely upon distinctions drawn according to race.\\
6 & 86 & Over the years, this Court has consistently repudiated "distinctions between citizens solely because of their ancestry" as being "odious to a free people whose institutions are founded upon the doctrine of equality."\\
8 & 88 & At the very least, the Equal Protection Clause demands that racial classifications, especially suspect in criminal statutes, be subjected to the "most rigid scrutiny," Korematsu v. United States, 323 U.S. 214, 216 (1944), and, if they are ever to be upheld, they must be shown to be necessary to the accomplishment of some permissible state objective, independent of the racial discrimination which it was the object of the Fourteenth Amendment to eliminate.\\
9 & 99 & II.These statutes also deprive the Lovings of liberty without due process of law in violation of the Due Process Clause of the Fourteenth Amendment.\\
5 & 100 & The freedom to marry has long been recognized as one of the vital personal rights essential to the orderly pursuit of happiness by free men.\\
3 & 101 & Marriage is one of the "basic civil rights of man," fundamental to our very existence and survival.\\
1 & 105 & The Fourteenth Amendment requires that the freedom of choice to marry not be restricted by invidious racial discriminations.\\
2 & 106 & Under our Constitution, the freedom to marry, or not marry, a person of another race resides with the individual and cannot be infringed by the State.\\
4 & 107 & These convictions must be reversed.\\ \\
\end{tabular}

\end{table*}

\begin{table*}
    \tiny 
    \begin{tabular}{c c | p{14.0cm}}
    \multicolumn{3}{c}{\textbf{Meyer v. Neb. (81 sentences in original)}} \\
Rank & Position & Extracted sentence \\ \hline
7 & 23 & The enactment of such a statute comes reasonably within the police power of the state.\\
3 & 36 & Without doubt, it denotes not merely freedom from bodily restraint but also the right of the individual to contract, to engage in any of the common occupations of life, to acquire useful knowledge, to marry, establish a home and bring up children, to worship God according to the dictates of his own conscience, and generally to enjoy those privileges long recognized at common law as essential to the orderly pursuit of happiness by free men.\\
1 & 38 & The established doctrine is that this liberty may not be interfered with, under the guise of protecting the public interest, by legislative action which is arbitrary or without reasonable relation to some purpose within the competency of the State to effect.\\
2 & 39 & Determination by the legislature of what constitutes proper exercise of police power is not final or conclusive but is subject to supervision by the courts.\\
8 & 41 & The American people have always regarded education and acquisition of knowledge as matters of supreme importance which should be diligently promoted.\\
9 & 43 & Corresponding to the right of control, it is the natural duty of the parent to give his children education suitable to their station in life; and nearly all the States, including Nebraska, enforce this obligation by compulsory laws.\\
4 & 73 & Adams v. Tanner, supra, p. 594, pointed out that mere abuse incident to an occupation ordinarily useful is not enough to justify its abolition, although regulation may be entirely proper.\\
6 & 74 & No emergency has arisen which renders knowledge by a child of some language other than English so clearly harmful as to justify its inhibition with the consequent infringement of rights long freely enjoyed.\\
10 & 76 & As the statute undertakes to interfere only with teaching which involves a modern language, leaving complete freedom as to other matters, there seems no adequate foundation for the suggestion that the pupose was to protect the child's health by limiting his mental activities.\\
5 & 77 & It is well known that proficiency in a foreign language seldom comes to one not instructed at an early age, and experience shows that this is not injurious to the health, morals or understanding of the ordinary child.\\ \\
\end{tabular}

    \begin{tabular}{c c | p{14.0cm}}
    \multicolumn{3}{c}{\textbf{United States v. Carolene Products Co. (94 sentences in original)}} \\
Rank & Position & Extracted sentence \\ \hline
1 & 17 & The power to regulate commerce is the power "to prescribe the rule by which commerce is to be governed," Gibbons v. Ogden, 9 Wheat. 1, 196, and extends to the prohibition of shipments in such commerce.\\
2 & 19 & The power "is complete in itself, may be exercised to its utmost extent and acknowledges no limitations other than are prescribed by the Constitution."\\
3 & 21 & Hence Congress is free to exclude from interstate commerce articles whose use in the states for which they are destined it may reasonably conceive to be injurious to the public health, morals or welfare, Reid v. Colorado, supra;Lottery Case, supra; Hipolite Egg Co. v. United States, 220 U.S. 45;Hope v. United States, supra, or which contravene the policy of the state of their destination.\\
4 & 23 & Such regulation is not a forbidden invasion of state power either because its motive or its consequence is to restrict the use of articles of commerce within the states of destination, and is not prohibited unless by the due process clause of the Fifth Amendment.\\
6 & 24 & And it is no objection to the exertion of the power to regulate interstate commerce that its exercise is attended by the same incidents which attend the exercise of the police power of the states.\\
8 & 71 & Even in the absence of such aids the existence of facts supporting the legislative judgment is to be presumed, for regulatory legislation affecting ordinary commercial transactions is not to be pronounced unconstitutional unless in the light of the facts made known or generally assumed it is of such a character as to preclude the assumption that it rests upon some rational basis within the knowledge and experience of the legislators.\\
9 & 72 & 4There may be narrower scope for operation of the presumption of constitutionality when legislation appears on its face to be within a specific prohibition of the Constitution, such as those of the first ten amendments, which are deemed equally specific when held to be embraced within the Fourteenth.\\
5 & 80 & Where the existence of a rational basis for legislation whose constitutionality is attacked depends upon facts beyond the sphere of judicial notice, such facts may properly be made the subject of judicial inquiry, Borden's Farm Products Co. v. Baldwin, 293 U.S. 194, and the constitutionality of a statute predicated upon the existence of a particular state of facts may be challenged by showing to the court that those facts have ceased to exist.\\
7 & 82 & Similarly we recognize that the constitutionality of a statute, valid on its face, may be assailed by proof of facts tending to show that the statute as applied to a particular article is without support in reason because the article, although within the prohibited class, is so different from others of the class as to be without the reason for the prohibition, Railroad Retirement Board v. Alton R. Co., 295 U.S. 330, 349, 351, 352; see Whitney v. California, 274 U.S. 357, 379; cf. Morf v. Bingaman, 298 U.S. 407, 413, though the effect of such proof depends on the relevant circumstances of each case, as for example the administrative difficulty of excluding the article from the regulated class.\\
10 & 84 & But by their very nature such inquiries, where the legislative judgment is drawn in question, must be restricted to the issue whether any state of facts either known or which could reasonably be assumed affords support for it.\\ \\
\end{tabular}

    \begin{tabular}{c c | p{14.0cm}}
    \multicolumn{3}{c}{\textbf{W. Va. State Bd. of Educ. v. Barnette (198 sentences in original)}} \\
Rank & Position & Extracted sentence \\ \hline
3 & 4 & 1§ 1734, West Virginia Code (1941 Supp.):"In all public, private, parochial and denominational schools located within this state there shall be given regular courses of instruction in history of the United States, in civics, and in the constitutions of the United States and of the State of West Virginia, for the purpose of teaching, fostering and perpetuating the ideals, principles and spirit of Americanism, and increasing the knowledge of the organization and machinery of the government of the United States and of the state of West Virginia.\\
6 & 5 & The state board of education shall, with the advice of the state superintendent of schools, prescribe the courses of study covering these subjects for the public elementary and grammar schools, public high schools and state normal schools.\\
4 & 6 & It shall be the duty of the officials or boards having authority over the respective private, parochial and denominational schools to prescribe courses of study for the schools under their control and supervision similar to those required for the public schools."\\
10 & 8 & 2The text is as follows:"WHEREAS, The West Virginia State Board of Education holds in highest regard those rights and privileges guaranteed by the Bill of Rights in the Constitution of the United States of America and in the Constitution of West Virginia, specifically, the first amendment to the Constitution of the United States as restated in the fourteenth amendment to the same document and in the guarantee of religious freedom in Article III of the Constitution of this State, and"WHEREAS, The West Virginia State Board of Education honors the broad principle that one's convictions about the ultimate mystery of the universe and man's relation to it is placed beyond the reach of law; that the propagation of belief is protected whether in church or chapel, mosque or synagogue, tabernacle or meeting house; that the Constitutions of the United States and of the State of West Virginia assure generous immunity to the individual from imposition of penalty for offending, in the course of his own religious activities, the religious views of others, be they a minority or those who are dominant in the government, but"WHEREAS, The West Virginia State Board of Education recognizes that the manifold character of man's relations may bring his conception of religious duty into conflict with the secular interests of his fellowman; that conscientious scruples have not in the course of the long struggle for religious toleration relieved the individual from obedience to the general law not aimed at the promotion or restriction of the religious beliefs; that the mere possession of convictions which contradict the relevant concerns of political society does not relieve the citizen from the discharge of political responsibility, and"WHEREAS, The West Virginia State Board of Education holds that national unity is the basis of national security; that the flag of our Nation is the symbol of our National Unity transcending all internal differences, however large within the framework of the Constitution; that the Flag is the symbol of the Nation's power; that emblem of freedom in its truest, best sense; that it signifies government resting on the consent of the governed, liberty regulated by law, protection of the weak against the strong, security against the exercise of arbitrary power, and absolute safety for free institutions against foreign aggression, and"WHEREAS, The West Virginia State Board of Education maintains that the public schools, established by the legislature of the State of West Virginia under the authority of the Constitution of the State of West Virginia and supported by taxes imposed by legally constituted measures, are dealing with the formative period in the development in citizenship that the Flag is an allowable portion of the program of schools thus publicly supported."Therefore, be it RESOLVED, That the West Virginia Board of Education does hereby recognize and order that the commonly accepted salute to the Flag of the United States -the right hand is placed upon the breast and the following pledge repeated in unison: 'I pledge allegiance to the Flag of the United States of America and to the Republic for which it stands; one Nation, indivisible, with liberty and justice for all' -now becomes a regular part of the program of activities in the public schools, supported in whole or in part by public funds, and that all teachers as defined by law in West Virginia and pupils in such schools shall be required to participate in the salute, honoring the Nation represented by the Flag; provided, however, that refusal to salute the Flag be regarded as an act of insubordination, and shall be dealt with accordingly."\\
1 & 22 & 5§ 1851 (1), West Virginia Code (1941 Supp.):"If a child be dismissed, suspended, or expelled from school because of refusal of such child to meet the legal and lawful requirements of the school and the established regulations of the county and/or state board of education, further admission of the child to school shall be refused until such requirements and regulations be complied with.\\
2 & 23 & Any such child shall be treated as being unlawfully absent from school during the time he refuses to comply with such requirements and regulations, and any person having legal or actual control of such child shall be liable to prosecution under the provisions of this article for the absence of such child from school."\\
7 & 24 & and may be proceeded against as a delinquent.\\
8 & 25 & 6§ 4904 (4), West Virginia Code (1941 Supp.).\\
5 & 26 & His parents or guardians are liable to prosecution, 7See Note 5, supra. and if convicted are subject to fine not exceeding \$ 50 and jail term not exceeding thirty days.\\
9 & 27 & 8§§ 1847, 1851, West Virginia Code (1941 Supp.).Appellees, citizens of the United States and of West Virginia, brought suit in the United States District Court for themselves and others similarly situated asking its injunction to restrain enforcement of these laws and regulations against Jehovah's Witnesses.\\ \\
\end{tabular}

\end{table*}

\begin{table*}

\tiny
    \begin{tabular}{c c | p{14.0cm}}
    \multicolumn{3}{c}{\textbf{Planned Parenthood v. Casey (862 sentences in original)}} \\
Rank & Position & Extracted sentence \\ \hline
3 & 50 & Thus all fundamental rights comprised within the term liberty are protected by the Federal Constitution from invasion by the States."\\
4 & 55 & We have held that the Due Process Clause of the Fourteenth Amendment incorporates most of the Bill of Rights against the States.\\
6 & 57 & It is tempting, as a means of curbing the discretion of federal judges, to suppose that liberty encompasses no more than those rights already guaranteed to the individual against federal interference by the express provisions of the first eight Amendments to the Constitution.\\
2 & 93 & It is conventional constitutional doctrine that where reasonable people disagree the government can adopt one position or the other.\\
7 & 95 & That theorem, however, assumes a state of affairs in which the choice does not intrude upon a protected liberty.\\
8 & 98 & Our law affords constitutional protection to personal decisions relating to marriage, procreation, contraception, family relationships, child rearing, and education.\\
1 & 100 & Our cases recognize "the right of the individual, married or single, to be free from unwarranted governmental intrusion into matters so fundamentally affecting a person as the decision whether to bear or beget a child."\\
9 & 132 & Indeed, the very concept of the rule of law underlying our own Constitution requires such continuity over time that a respect for precedent is, by definition, indispensable.\\
10 & 134 & At the other extreme, a different necessity would make itself felt if a prior judicial ruling should come to be seen so clearly as error that its enforcement was for that very reason doomed.\\
5 & 135 & Even when the decision to overrule a prior case is not, as in the rare, latter instance, virtually foreordained, it is common wisdom that the rule of stare decisis is not an "inexorable command," and certainly it is not such in every constitutional case, see Burnet v. Coronado Oil \& Gas Co., 285 U.S. 393, 405-411, 76 L. Ed. 815, 52 S. Ct. 443 (1932) (Brandeis, J., dissenting).\\ \\
\end{tabular}

    \begin{tabular}{c c | p{14.0cm}}
    \multicolumn{3}{c}{\textbf{San Antonio Indep. Sch. Dist. v. Rodriguez (649 sentences in original)}} \\
Rank & Position & Extracted sentence \\ \hline
2 & 161 & Texas virtually concedes that its historically rooted dual system of financing education could not withstand the strict judicial scrutiny that this Court has found appropriate in reviewing legislative judgments that interfere with fundamental constitutional rights 39E.g., Police Dept. of Chicago v. Mosley, 408 U.S. 92 (1972); Dunn v. Blumstein, 405 U.S. 330 (1972); Shapiro v. Thompson, 394 U.S. 618 (1969). or that involve suspect classifications.\\
1 & 163 & If, as previous decisions have indicated, strict scrutiny means that the State's system is not entitled to the usual presumption of validity, that the State rather than the complainants must carry a "heavy burden of justification," that the State must demonstrate that its educational system has been structured with "precision," and is "tailored" narrowly to serve legitimate objectives and that it has selected the "less drastic means" for effectuating its objectives 41See Dunn v. Blumstein, supra, at 343, and the cases collected therein.\\
5 & 169 & We must decide, first, whether the Texas system of financing public education operates to the disadvantage of some suspect class or impinges upon a fundamental right explicitly or implicitly protected by the Constitution, thereby requiring strict judicial scrutiny.\\
7 & 170 & If so, the judgment of the District Court should be affirmed.\\
4 & 171 & If not, the Texas scheme must still be examined to determine whether it rationally furthers some legitimate, articulated state purpose and therefore does not constitute an invidious discrimination in violation of the Equal Protection Clause of the Fourteenth Amendment.\\
3 & 183 & This approach largely ignores the hard threshold questions, including whether it makes a difference for purposes of consideration under the Constitution that the class of disadvantaged "poor" cannot be identified or defined in customary equal protection terms, and whether the relative -rather than absolute -nature of the asserted deprivation is of significant consequence.\\
6 & 184 & Before a State's laws and the justifications for the classifications they create are subjected to strict judicial scrutiny, we think these threshold considerations must be analyzed more closely than they were in the court below.\\ \\
\end{tabular}

    \begin{tabular}{c c | p{14.0cm}}
    \multicolumn{3}{c}{\textbf{Olmstead v. United States (177 sentences in original)}} \\
Rank & Position & Extracted sentence \\ \hline
1 & 35 & The Fourth Amendment provides -"The right of the people to be secure in their persons, houses, papers, and effects against unreasonable searches and seizures shall not be violated; and no warrants shall issue but upon probable cause, supported by oath or affirmation and particularly describing the place to be searched and the persons or things to be seized."\\
2 & 36 & And the Fifth: "No person… shall be compelled, in any criminal case, to be a witness against himself."\\
5 & 40 & The fifth section of the Act of June 22, 1874, provided that in cases not criminal under the revenue laws, the United States Attorney, whenever he thought an invoice, belonging to the defendant, would tend to prove any allegation made by the United States, might by a written motion describing the invoice and setting forth the allegation which he expected to prove, secure a notice from the court to the defendant to produce the invoice, and if the defendant refused to produce it, the allegations stated in the motion should be taken as confessed, but if produced, the United States Attorney should be permitted, under the direction of the court, to make an examination of the invoice, and might offer the same in evidence.\\
3 & 50 & It is our opinion, therefore, that a compulsory production of a man's private papers to establish a criminal charge against him, or to forfeit his property, is within the scope of the Fourth Amendment to the Constitution, in all cases in which a search and seizure would be; because it is a material ingredient, and effects the sole object and purpose of search and seizure."\\
10 & 126 & This Court in Carroll v. United States, 267 U.S. 132, 149, declared: "The Fourth Amendment is to be construed in the light of what was deemed an unreasonable search and seizure when it was adopted and in a manner which will conserve public interests as well as the interests and rights of individual citizens."\\
9 & 142 & The rules of evidence in criminal cases in courts of the United States sitting there, consequently are those of the common law.\\
6 & 144 & The common law rule is that the admissibility of evidence is not affected by the illegality of the means by which it was obtained.\\
8 & 145 & Professor Greenleaf in his work on evidence, vol. 1, 12th ed., by Redfield, § 254(a) says: "It may be mentioned in this place, that though papers and other subjects of evidence may have been illegally taken from the possession of the party against whom they are offered, or otherwise unlawfully obtained, this is no valid objection to their admissibility, if they are pertinent to the issue.\\
7 & 146 & The court will not take notice how they were obtained, whether lawfully or unlawfully, nor will it form an issue, to determine that question."\\
4 & 165 & The statute of Washington, adopted in 1909, provides (Remington Compiled Statutes, 1922, § 2656-18) that: "Every person… who shall intercept, read or in any manner interrupt or delay the sending of a message over any telegraph or telephone line… shall be guilty of a misdemeanor."\\ \\
\end{tabular}

\vspace{20cm}
\end{table*}

\end{document}